\documentclass[letterpaper]{article}
\usepackage{aaai}
\usepackage{times}
\usepackage{helvet}
\usepackage{courier}
\usepackage{graphicx}
\usepackage{amsmath}
\usepackage{amssymb}
\usepackage{multirow}
\begin{document}

\title{Spatial-Aware Non-Local Attention for Fashion Landmark Detection}

\author{Yixin Li \\Peking University\\liyixin@pku.edu.cn \And Shengqin Tang \\Xi’an Jiaotong University\\tsq292978891@stu.xjtu.edu.cn \And Yun Ye \\JD.com\\yeyun@jd.com \And Jinwen Ma \\Peking University\\jwma@math.pku.edu.cn
}

\maketitle

\begin{abstract}

Fashion landmark detection is a challenging task even using the current deep learning techniques, due to the large variation and non-rigid deformation of clothes. In order to tackle these problems, we propose Spatial-Aware Non-Local (SANL) block, an attentive module in deep neural network which can utilize spatial information while capturing global dependency. Actually, the SANL block is constructed from the non-local block in the residual manner which can learn the spatial related representation by taking a spatial attention map from Grad-CAM. We then establish our fashion landmark detection framework on feature pyramid network, equipped with four SANL blocks in the backbone. It is demonstrated by the experimental results on two large-scale fashion datasets that our proposed fashion landmark detection approach with the SANL blocks outperforms the current state-of-the-art methods considerably. Some supplementary experiments on fine-grained image classification also show the effectiveness of the proposed SANL block.

\end{abstract}

\section{Introduction}

Nowadays, fashion is a trillion dollar industry \cite{latent_look}, and understanding the fashion images has become a hot topic in computer vision. Fashion landmark detection is one of the key issues on fashion image understanding, which is to localize the functional regions like sleeves and hems of clothes. Fashion landmarks can be used for improving the performance of applications such as clothes retrieval \cite{fashion_landmark} and attribute prediction \cite{fashion_landmark,fashion_grammar}. 

Human pose estimation \cite{hourglass} is a similar task, while fashion landmark detection is more challenging due to the various appearances and non-rigid deformation of clothes. To overcome these difficulties, the model requires a large receptive field to determine whether the target location is a landmark and which type of landmark it is. For instance, it is hard to tell left and right sleeves just based on a small region, for the reason that the image features are nearly the same if we ignore the hands of the model or the whole clothes. 

Modern deep learning techniques tend to increase receptive fields by stacking convolution layers \cite{vgg,resnet}. Recently, non-local neural network \cite{non-local_network} provides a new strategy to require a large receptive fields and capture global dependency. In the non-local operation, the output response at a position is computed by features of all positions on the input feature map, and the weights and transform functions are learned in the end-to-end manner. 

We notice that there is prior knowledge that could help the non-local block to learn a better representation. Take fashion landmark detection as an example, we know that most landmarks (at least the visible ones) are on the cloth area. Visual attention mechanism \cite{attentional_detection,show_attention_tell,residual_attention,ra_cnn} has proven to be a useful technique in various computer vision tasks. It allows the feature learning to focus on a significant location and enhance the representation discrimination on that location. So if we encode this prior knowledge into the original non-local block as an attention mechanism, it may enhance the feature representations of the target area.

Inspired by visual attention mechanism \cite{residual_attention} and non-local block \cite{non-local_network}, we propose Spatial-Aware Non-Local (SANL) block, a basic building block which can be easily plugged into any modern deep neural networks. Based on the original non-local block, our SANL block encodes prior knowledge by taking a spatial attention map in the calculation of the similarity matrix. So, high-level semantic information can be passing into the blocks directly instead of travelling all the way from the top of the deep neural network.

\begin{figure*}
    \centering
    \includegraphics[width=1.0\textwidth]{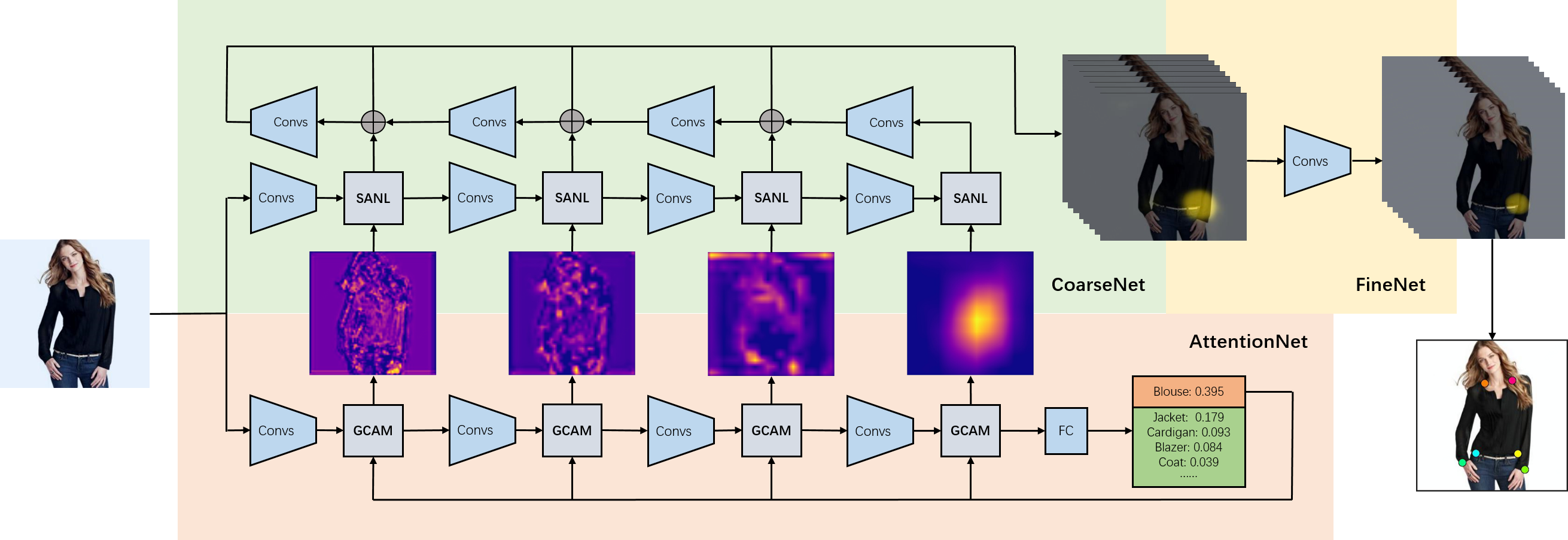}
    \caption{\textbf{The framework of our proposed fashion landmark detection approach.} AttentionNet (orange) is a classification network pretrained on category annotations and provides spatial attention maps by Grad-CAM. CoarseNet (green) is a feature pyramid network with our proposed SANL blocks. FineNet (yellow) is a top-down bottom-up unit which refines the coarse map generated by CoarseNet to the fine map. The final fashion landmark locations are computed from the fine map.}
    \label{fig:pipeline}
\end{figure*}

Our fashion landmark detection framework is built on SANL blocks. We give a feasible way of deploying SANL blocks without using extra data other than DeepFasion \cite{deepfashion}, which is adopting Grad-CAM \cite{grad-cam} to generate spatial attention maps. A classification network is trained under the supervision of category annotations to further exploit the semantic regions. To be specific, the attention maps on the certain feature layers are computed by Grad-CAM and fed into our SANL blocks as spatial attention maps. As shown in Figure. \ref{fig:pipeline}, the whole model is based on feature pyramid network \cite{FPN} and designed in the coarse-to-fine manner.

Our main contribution of this paper is three-fold: (1) We propose Spatial-Aware Non-Local (SANL) block which encodes the prior knowledge by feeding a spatial attention map into the non-local block. (2) We demonstrate a feasible way to deploy our proposed SANL block in deep neural network, which computes the input spatial attention maps of a pretrained network by Grad-CAM. (3) We establish the fashion landmark detection model based on our proposed SANL blocks, and outperform the current state-of-the-art approaches without using extra dataset.

The rest of this paper is organized as follows: We first summarize the related work on visual attention mechanism and fashion image understanding. Then the Spatial-Aware Non-Local block and fashion landmark detection approach are presented in details. Experiments along with implementation details and ablation study are given next. Finally we conclude this paper in the last section.

\section{Related Work}

Since fashion landmark detection is a key topic of fashion image understanding and our proposed Spatial-Aware Non-Local block performs in the visual attention mechanism, we briefly introduce the recent work on fashion image understanding and visual attention mechanism in this section.

\subsection{Fashion Image Understanding}

Powering by the large-scale fashion datasets \cite{deepfashion,fashion_landmark,ATR}, deep learning based models for fashion related tasks have boomed rapidly in recent years. The key problems of fashion image understanding includes recognition \cite{cloth_attributes,fashion_grammar}, retrieval \cite{wheretobuy,cross-domain_retrieval,deepfashion}, recommendation \cite{capsule_recom,lstm_recom}, generation \cite{deformable,pose_guided}, and landmark detection \cite{fashion_landmark,DLAN,fashion_grammar}.

Fashion landmark detection is a rather new topic in fashion understanding, so there is not much prior work we can refer to. In fact, the related approaches can be roughly divided into two category: coordinate based \cite{deepfashion,fashion_landmark,DLAN} and heatmap based \cite{fashion_grammar}. FashionNet \cite{deepfashion} is based on VGG-16 and learns the coordinate and visibility of the landmark directly. FLD \cite{fashion_landmark} utilizes auto-routing mechanism to reduce the large variations in fashion images. DLAN \cite{DLAN} uses selective dilated convolutions for handling scale discrepancies, and a hierarchical recurrent spatial transformer for handling background clutters. And the most recent work \cite{fashion_grammar} constructed the relation between fashion landmarks which is called ``fashion grammar'' and proposed bidirectional convolutional recurrent neural network (BCRNN) to learn the landmark heatmaps.

\subsection{Visual Attention Mechanism}

Attention mechanism has been widely applied in natural language processing, especially machine translation \cite{machine_translation,self-attention}. Inspired by self-attention \cite{self-attention} for machine translation, which computes the response by attending to all positions on a sequence, non-local neural network \cite{non-local_network} introduces self-attention into computer vision by performing a similar computation on feature maps. So, non-local block \cite{non-local_network} can be viewed as a particular form of self-attention \cite{self-attention}. 

Visual attention is rather common in human perception, for example, bright color can easily draw our attention and we can locate a cat just by a glance. Attention mechanism has been introduced to deep neural networks and is widely applied to various computer vision tasks, such as image classification \cite{multiple_attention,residual_attention,ra_cnn}, object detection \cite{attentional_detection,attentive_contexts}, image captioning \cite{show_attention_tell,sca_cnn,butd_attention} and visual question answering \cite{hierarchical_qa,butd_attention}.

\section{Methodology}

We first introduce the Spatial-Aware Non-Local (SANL) attention mechanism in detail. Then we describe our fashion landmark detection approach built on SANL blocks.

\begin{figure}
    \centering
    \includegraphics[width=0.45\textwidth]{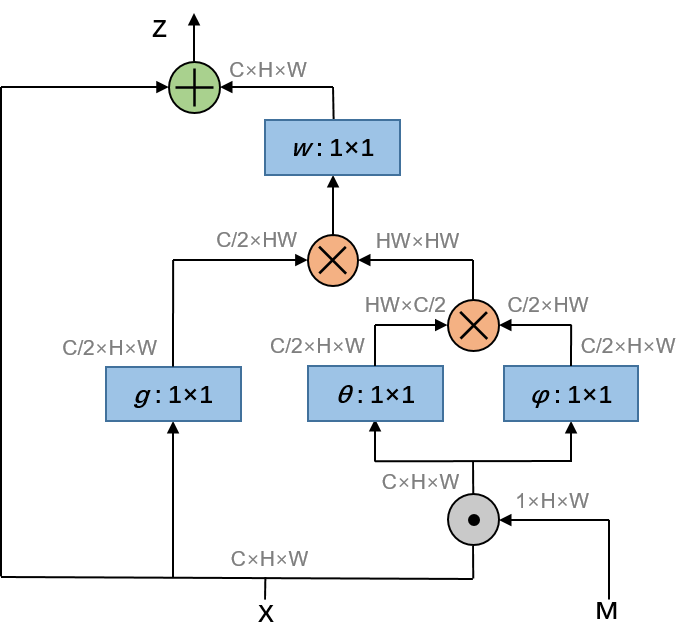}
    \caption{\textbf{The Spatial-Aware Non-Local (SANL) block.} A spatial attention map $M$ is added on the input feature map $X$ by Hadamard multiplication ($\cdot$) before the computation of similarity matrix. Then the similarity matrix and the response of value are combined by matrix multiplication ($\times$). The whole block is learned in the residual manner.}
    \label{fig:SANL}
\end{figure}

\subsection{Spatial-Aware Non-Local Attention}

Non-local means method \cite{non-local_denoising} is a classical computer vision filtering algorithm where each location in the filtered image is determined by all the pixels in the input image. Inspired by this work, non-local network \cite{non-local_network} is proposed to perform a similar computation in deep learning architecture as a special case of self-attention \cite{self-attention}. The non-local block computes the weighted sum of the response at each position on the feature map which lead to a receptive field same as the feature map size.

\textbf{The original non-local block.} The formulation of the original non-local block is borrowed from self-attention \cite{self-attention} for machine translation. In a self-attention block, the similarity of every pair of 'query' and 'key' is computed, then normalized by Softmax function, and the final attention map is the weighted sum of the normalized similarity and 'value'. In a non-local block, 'query', 'key' and 'value' are all input feature map. The non-local operation can be denoted as:

\begin{equation}
Y_i = \frac{1}{C(X)}\sum_j{f(X_i,X_j)g(X_j)}
\end{equation}
\\
where $i$, $j$ is location index, $f(\cdot,\cdot)$ is the similarity function, $g(\cdot)$ computes the response of input feature map $X$, and $C(X)$ is the normalize factor. 

After the non-local operation, the transform function $w$ is applied on the output $Y$, then added on the the original input feature map $X$. That is, the whole non-local block is learned in the residual learning manner. 

\textbf{Spatial-Aware Non-Local (SANL) block.} The receptive field of the original non-local block is the whole feature map, but the weights of each location are learned in an end-to-end manner which might be a long way for the gradients to pass backward. Under the observation that only the cloth area in the image matters the most on fashion landmark detection task, encoding this prior knowledge might help the model focus on the target area and learn a better representation. So the intuition of the Spatial-Aware Non-Local (SANL) is that encode the spatial information in the similarity function to help the non-local block learn a better weights over the spatial locations.

Given a spatial attention map $M$, the formulation of our proposed SANL operation is :

\begin{equation}
y_i = \frac{1}{C(X)}\sum_j{f(h(X_i, M),h(X_j, M))g(X_j)}
\end{equation}
\\
where $h(\cdot,\cdot)$ is the transform function that applies spatial attention map $M$ on the input feature map $X$.

\textbf{Implementation details of the SANL blocks.} In our paper, we choose matrix dot-product as our similarity function, and Hadamard product for function $h$. So the similarity function can be written as:

\begin{equation}
f(X_i,X_j) = \theta[X_i*(1+M_i)]^T\phi[X_j*(1+M_j)]
\end{equation}

Inherited from \cite{non-local_network}, the transform functions $\theta(\cdot)$, $\phi(\cdot)$ and $g(\cdot)$ are all $1\times1$ convolution layers. The normalization function $C(X)$ is equal to the input feature map size $N$. The whole pipeline of our proposed SANL block can be found in Fig.\ref{fig:SANL}.

The spatial attention map $M$ can be different response of spatial information, such as parsing and segmentation map. To avoid using extra data, we adopt Gradient-weighted Class Activation Mapping (Grad-CAM) \cite{grad-cam} to compute the attention map $M$ from the model pretrained on category annotation. The details about the SANL blocks in our fashion landmark detection model can be found in the next subsection.

\subsection{Fashion Landmark Detection Framework}

Fashion landmark detection is a similar task to human keypoint detection. Recent trend of human keypoint detection is to predict separated heatmaps of the corresponding keypoints \cite{hourglass}, for the reason that directly regression of keypoints or landmarks coordinates is a rather difficult task for a very deep neural network to learn. So we adopt the same method to predict a heatmap with 8 channels for 8 different fashion landmarks. 

\textbf{Base model.} Feature pyramid network (FPN) \cite{FPN} builds a in-network feature pyramid architecture with single scale input. It achieves great performance in object detection and in the mean time, it is highly efficient. Mask R-CNN \cite{mask_rcnn} takes FPN as backbone and achieves state-of-the-art on object detection and instance segmentation, even on human keypoint detection. Also, Stacked Hourglass \cite{hourglass}, the widely used landmark detection approach, has a very similar feature hierarchy structure.

In a feature pyramid network, a top-down structure is connected with the bottom-up backbone by lateral connections so that the high-level semantic features can be passed to the low-level feature maps. And multi-scale predictions are made from the top-down architecture at all scales. For more details about FPN, we refer the FPN paper \cite{FPN}.

Since fashion landmark detection is also a highly spatial related task, we choose FPN as our base model. We take ResNet-101 \cite{resnet} as our backbone network and inherited the lateral connections and top-down network structure from the FPN paper \cite{FPN}. Multi-scale predictions are merged by concatenation and followed by an Atrous Spatial Pyramid Pooling (ASPP) block \cite{deeplabv3} to predict a stride 4 heatmap for each landmark. The whole network is in a fully convolutional way.

\textbf{Gradient-weighted Class Activation Mapping (Grad-CAM).} Grad-CAM is a visualization method for deep neural network. Unlike the global average pooling layer is essential in CAM \cite{cam}, Grad-CAM can be applied to any network structure by computing the gradients of the feature maps as weights of the feature map channels. The details can be found in the paper \cite{grad-cam}.

We pretrain a ResNet-18 network on DeepFashion-C category annotations \cite{deepfashion} which contains 50 common categories of clothes. And the spatial attention maps on the last feature maps of each stride are calculated by Grad-CAM on the predicted category. As shown in \ref{fig:pipeline}, the activation map from low-level feature maps contains the spatial information of low-level features, while high-level activation map shows more semantic spatial information (the cloth whereabouts).

\textbf{SANL blocks for landmark detection.} In the backbone ResNet-101 of our fashion landmark detection model, four SANL blocks are placed after every last convolution layers at stride 4, 8, 16 and 32, that is conv2\_3, conv3\_4, conv4\_23 and conv5\_3. 

We take above Grad-CAM maps as the spatial attention maps of our SANL blocks. The first way is to use high-level activation map only. And the other way is that the attention map is generated from the same stride. That is, SANL block at low-level takes low-level activation map as spatial attention and high-level SANL block takes high-level activation map. The comparison of these two approaches can be found in the ablation study of this paper.

\begin{figure}
    \centering
    \includegraphics[width=0.40\textwidth]{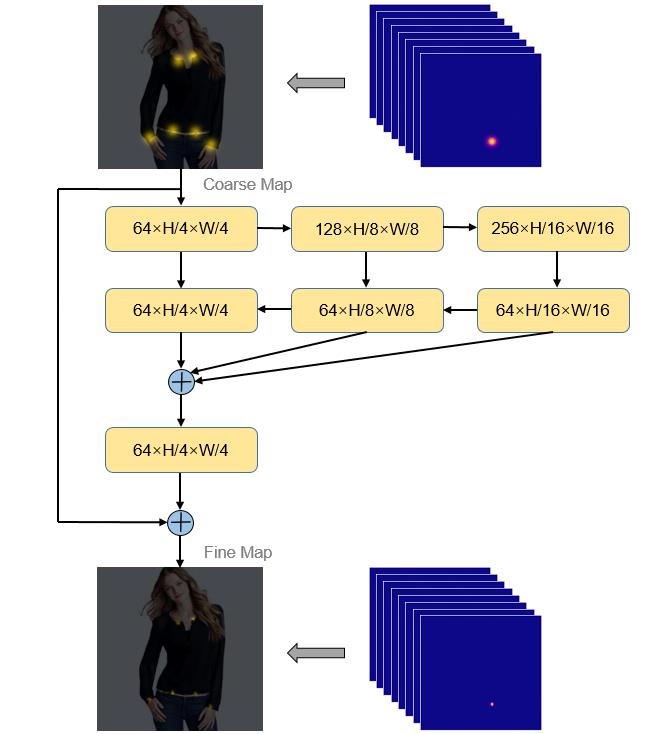}
    \caption{\textbf{The network structure of our proposed FineNet.} The FineNet is based on top-down bottom-up structure and plays the role of a refine unit. The whole network can be trained in end-to-end under the supervision of both coarse ground truth map and fine ground truth map.}
    \label{fig:C2F}
\end{figure}

\textbf{Coarse-to-fine architecture.} The ground truth heatmap is generated by adopting Gaussian filter on an one-hot map at landmark location. The parameter $\sigma$ of Gaussian kernel controls the radius of activation around landmark. If $\sigma$ is too small, the network tends to learn a near-zero map due to the imbalance of positive (non-zero) and negative (zero) data. On the other hand, the large prediction area will lead to accuracy loss if $\sigma$ is too large.

To tackle the contradiction of the parameter $\sigma$, we construct a coarse-to-fine architecture. The base model we called CoarseNet is trained on large $\sigma$ ground truth. FineNet takes the output coarse prediction as input, produces fine prediction which is under supervision of small $\sigma$ ground truth. After CoarseNet is pretrained, we fine-tune the whole network under both coarse and fine ground truth map.

As shown in Figure \ref{fig:C2F}, we design FineNet in a bottom-up top-down manner. The multi-scale feature maps are merged by adding up directly. And the whole module learns the residual of coarse map and fine map.

Instead of learning a highly imbalanced ground truth map, this cascaded structure allows the CoarseNet to learn a rough location of landmarks first, and FineNet to learn the exact positions. And FineNet play a role as a boundary refinement module which is similar to the ones in Global Convolution Network (GCN) \cite{GCN}.

\textbf{Loss function.} Instead of using mean square error loss \cite{hourglass}, we adopt weighted binary cross-entropy loss for a better and faster optimization result under the thought of "decouple" in \cite{mask_rcnn}. 

We also find that instead of calculating the average of loss over all the feature maps, only passing the gradient of the feature maps with visible landmarks brings a significant performance gain. The loss function can be denoted as:

\begin{equation}
\begin{split}
L = \frac{1}{N}\sum_{i,j}\sum_{c\in C_{vis}}&{[\lambda_p t_{i,j}^clog(\sigma(x_{i,j}^c))}\\
&+(1-t_{i,j}^c)log(1-\sigma(x_{i,j}^c))]
\end{split}
\end{equation}
\\
where $t_{i,j}^c$ is ground truth activation of landmark $c$ on location $i,j$ and $x_{i,j}^c$ is the prediction. $\sigma(\cdot)$ is Sigmoid function. $\lambda_p$ is weight of positive samples. $C_{vis}$ is the visible landmark index set.

\section{Experiment}

In this section, we report the results of our proposed fashion landmark detection approach on DeepFashion-C \cite{deepfashion} and FLD \cite{fashion_landmark}. The implementation details and ablation study are also given in this section.

\begin{table*}[]
    \centering
    \begin{tabular}{c||c|c|c|c|c|c|c|c||c}
        Methods & L.Collar & R.Collar & L.Sleeve & R.Sleeve & L.Waist & R.Waist & L.Hem & R.Hem & Avg.  \\
        \hline
        FashionNet \cite{deepfashion} & .0854 & .0902 & .0973 & .0935 & .0854 & .0845 & .0812 & .0823 &.0872 \\ 
        DFA \cite{fashion_landmark} & .0628 & .0637 & .0658 & .0621 & .0726 & .0702 & .0658 & .0663 &.0660 \\ 
        DLAN \cite{DLAN} & .0570 & .0611 & .0672 & .0647 & .0703 & .0694 & .0624 & .0627 &.0643 \\
        AFGN \cite{fashion_grammar} & .0415 & .0404 & .0496 & .0449 & .0502 & .0523 & .0537 & .0551 &.0484 \\
        \hline
        Ours (VGG-16) & \textbf{.0277} & \textbf{.0282} & \textbf{.0391} & \textbf{.0394} & \textbf{.0297} & \textbf{.0299} & \textbf{.0395} & \textbf{.0401} & \textbf{.0342} \\
        Ours (ResNet-101) & \textbf{.0249} & \textbf{.0256} & \textbf{.0341} & \textbf{.0348} & \textbf{.0260} & \textbf{.0259} & \textbf{.0338} & \textbf{.0346} & \textbf{.0299} \\
    \end{tabular}
    \caption{The normalized errors on DeepFashion-C dataset.}
    \label{tab:DFC}
\end{table*}

\begin{table*}[]
    \centering
    \begin{tabular}{c||c|c|c|c|c|c|c|c||c}
        Methods & L.Collar & R.Collar & L.Sleeve & R.Sleeve & L.Waist & R.Waist & L.Hem & R.Hem & Avg.  \\
        \hline
        FashionNet \cite{deepfashion} & .0784 & .0803 & .0975 & .0923 & .0874 & .0821 & .0802 & .0893 & .0859 \\ 
        DFA \cite{fashion_landmark} & .048 & .048 & .091 & .089 & $-$ & $-$ & .071 & .072 &.068 \\ 
        DLAN \cite{DLAN} & .0531 & .0547 & .0705 & .0735 & .0752 & .0748 & .0693 & .0675 &.0672 \\
        AFGN \cite{fashion_grammar} & .0463 & .0471 & .0627 & .0614 & .0635 & .0692 & .0635 & .0527 & .0583 \\
        \hline
        Ours (VGG-16) & \textbf{.0296} & \textbf{.0298} & \textbf{.0489} & \textbf{.0471} & \textbf{.0402} & \textbf{.0413} & \textbf{.0546} & \textbf{.0580} & \textbf{.0437} \\
        Ours (ResNet-101) & \textbf{.0270} & \textbf{.0275} & \textbf{.0444} & \textbf{.0430} & \textbf{.0366} & \textbf{.0365} & \textbf{.0496} & \textbf{.0519} & \textbf{.0396} \\
    \end{tabular}
    \caption{The normalized errors on FLD dataset.}
    \label{tab:FLD}
\end{table*}

\subsection{Datasets}

We evaluate our fashion landmark detection model on two large-scale datasets: DeepFashion-C \cite{deepfashion} and FLD \cite{fashion_landmark}.

\textbf{DeepFashion-C}. DeepFashion: Category and Attribute Prediction Benchmark (DeepFashion-C) \cite{deepfashion} is a large-scale visual fashion understanding dataset. It contains near 300k images with category, attribute and landmark annotations. For fashion landmark detection, each image is labeled with up to 8 fashion landmarks along with their visibilities. The dataset is split into three subset, 210k for training, 40k for validation and 40k for testing.

\textbf{FLD}. Fashion Landmark dataset (FLD) \cite{fashion_landmark} is a subset of DeepFashion database. The annotations in FLD are carefully refined and divided into five subsets which are normal/medium/large poses and medium/large zoom-ins. FLD also splits the data into training, validation and test set. To be more specific, 83k images are for training while validation and test set has 20k images each.

\subsection{Implementation Details}

Our fashion landmark model is based on a ResNet-101 backbone with 4 SANL blocks, built with feature pyramid and coarse-to-fine architecture. A ResNet-18 is pretrained on the DeepFashion-C category annotations to generate the spatial attention maps of SANL blocks. All of our models are implemented in Python under the Pytorch 0.4.0 framework.

\textbf{Training.} We use the ResNet-101 backbone pretrained on ImageNet \cite{imagenet} for a faster convergence. And we use SGD with 5e-4 weight decay to optimize the loss function. The initial learning rate is set to $0.001$, and decreased by a factor of 0.1 every 10 training epochs. The input images are resized to $320\times320$ while keeping the original aspect ratio then normalized by the image mean and variance of ImageNet. We use the batch size of 32 images and train the CoarseNet for 32 epochs, then we fine-tune the whole model under both coarse map and fine map supervision for 5 epochs. The whole training procedure takes about 2 days on 4 Tesla V100 GPUs.

\textbf{Inference.} At inference time, the input image is resized to $320\times320$ in the same way as training. Our model generates a $8\times80\times80$ landmark heatmap for a single input image. Then we upsamle this heatmap to the same size as the input ($8\times320\times320$) by bilinear interpolation. Instead of finding the pixel with largest response, we calculate the centroid of the largest connected component on each channel of landmark heatmap to be the final detected landmark.

\subsection{Quantitative Results}

We compare our proposed fashion landmark detection approach with four latest work \cite{deepfashion,fashion_landmark,DLAN,fashion_grammar} on both DeepFasion-C and FLD dataset. 

The quantitative results are given by the Normalized Error (NE), which is the L2 distance between predicted landmark positions and ground truth landmark positions normalized by the edge of the image. Intuitively, take the input image size $320\times320$ for example, a 0.050 NE score means the detected landmarks are $320\times0.05=16$ pixels away from the ground truth landmarks on average. The NE scores of each fashion landmark and the average NE score are shown in Table. \ref{tab:DFC} and Table \ref{tab:FLD}. The lower NE score means the better performance of the model.

\begin{figure*}
    \centering
    \includegraphics[width=1.0\textwidth]{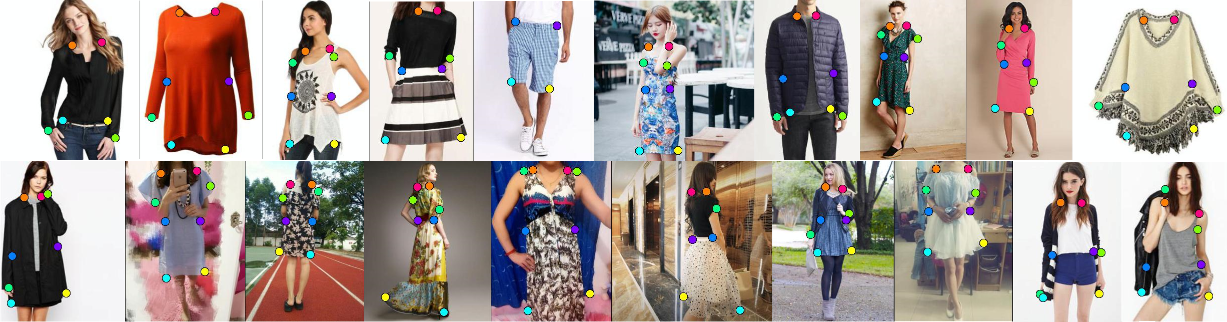}
    \caption{\textbf{Visualization results of our proposed fashion landmark detection approach.} Images on the first row are the results on DeepFashion-C test set, and results on FLD dataset are on the second row. Different landmarks are marked in different colors.}
    \label{fig:results}
\end{figure*}

As we can see in Table \ref{tab:DFC}, our proposed fashion landmark detection model achieves 0.0299 average NE score on DeepFashion-C dataset and 0.0396 on FLD. All the previous work \cite{deepfashion,fashion_landmark,DLAN,fashion_grammar} take VGG-16 \cite{vgg} as backbone structure. To be fair, we re-implement our model under VGG-16 backbone and keep the other network structure unchanged. Notice that we still outperform all the other approaches by 0.0342 on DeepFashion-C and 0.0437 on FLD. Some visualization results are shown in Figure \ref{fig:results}.

\subsection{Ablation Study}

We carry out some more experiments on our fashion landmark model to prove the effectiveness of our proposed module. All the experiments in this subsection are carried out on DeepFashion-C dataset and all the models are trained only on visible landmark maps under the same hyper-parameters.

\textbf{Effectiveness of the SANL blocks.} The baseline model is our proposed feature pyramid model and the other three models are considered: \\
$\bullet$ Base: our feature pyramid baseline model.\\
$\bullet$ NL: baseline model with non-local blocks.\\
$\bullet$ SANL(32): baseline model with the SANL blocks and the spatial maps are all from stride 32 Grad-CAM maps.\\
$\bullet$ SANL(all): baseline model with the SANL blocks and the spatial maps are from the corresponding strides.

Results are shown in Table \ref{tab:SANL}. Non-local blocks give a 0.0012 less average NE score than baseline. Our proposed SANL block further improves 0.0017 and 0.0021 by passing the spatial information into non-local block.

As for the comparison of different strides of spatial attention maps, we find that using the corresponding stride of the attention maps gives a slightly lower NE score than using high-level spatial attention map on all the SANL blocks. The explanation for this difference is that high-level and low-level non-local blocks play different roles in the model like normal convolution layers. The low-level non-local block learns basic image features while high-level non-local block learns semantic information. So feeding attention maps in the corresponding strides help the non-local blocks learn a better spatial related feature representation.

\begin{table}[]
    \centering
    \begin{tabular}{c||c c||c c}
        Methods & Params & Flops & NE & $\Delta$NE \\
        \hline
        Base & 1.00$\times$ & 1.00$\times$ & .0336 & $-$ \\
        NL & 1.28$\times$ & 1.12$\times$ & .0324 & .0012 \\
        SANL (32) & 1.28$\times$ & 1.13$\times$ & .0307 & .0029 \\
        SANL (all) & 1.28$\times$ & 1.13$\times$ & .0303 & .0033 \\
    \end{tabular}
    \caption{Ablation study on our proposed SANL attention.}
    \label{tab:SANL}
\end{table}

\textbf{Effectiveness of modules in our fashion landmark detection framework.} We also take our proposed feature pyramid model as baseline model and evaluate the performance of three modules on our baseline model:\\
$\bullet$ Vis: only trained on maps with visible landmarks.\\
$\bullet$ SANL: baseline model with the SANL blocks.\\
$\bullet$ C2F: baseline model with coarse-to-fine architecture.

From Table \ref{tab:model} we notice that only trained on visible maps brings a huge improvement on average NE score. Only ground truth maps with visible landmarks contribute to the final loss while the zero maps are ignored. This allows the training procedure concentrates more on visible landmarks and minimizes the effect of missing landmark annotations. 

Our proposed SANL block further improves the average NE score by 0.0033 with only 28\% more parameters and 13\% more computation cost. Our SANL block taking spatial attention maps into consideration, makes the network focus on the features on the most significant areas of fashion items. The well pretrained CoarseNet makes less room for FineNet to improve, so we only get a slightly fine-tuned version of CoarseNet from our coarse-to-fine architecture.

\begin{table}[]
    \centering
    \begin{tabular}{c c c||c c||c}
        Vis & SANL & C2F & Params & Flops & NE \\
        \hline
        & & & 1.00$\times$ & 1.00$\times$ & .0381\\
        $\checkmark$ & & & 1.00$\times$ & 1.00$\times$ & .0336 \\
        $\checkmark$ & $\checkmark$ & & 1.28$\times$ & 1.13$\times$ & .0303 \\
        $\checkmark$ & $\checkmark$ & $\checkmark$ & 1.29$\times$ & 1.60$\times$ & .0299 \\
    \end{tabular}
    \caption{Ablation study on different modules of our proposed fashion landmark detection approach.}
    \label{tab:model}
\end{table}

\subsection{Supplementary Experiments}

To further prove the effectiveness of our proposed SANL blocks, we conduct some more experiments on two fine-grained classification datasets: CUB-200 \cite{cub} and FoodNet \cite{foodnet}.

\textbf{Fine-grained classification.} Recognizing fine-grained categories such as bird species is a challenging task, due to the small discriminative regions. Our approach utilizes Grad-CAM to generate spatial attention maps, which makes the model pay more attention on the target regions, thus more sensitive to the discriminative features.

\textbf{Datasets.} Caltech-ucsd birds-200-2011 (CUB-200) \cite{cub} is a fine-grained dataset annotated by 200 bird species. The dataset is split into roughly equal train and test sets from the 11,788 images in total. ChineseFoodNet \cite{foodnet} is a large Chinese food collection with 185,628 images and 208 categories on Chinese dishes. Because the annotations of the test set are not disclosed in FoodNet, we report the results on the validation set in this paper.

\textbf{Base models.} We use Bilinear Convolutional Neural Network (BCNN) \cite{bcnn} and VGG19 \cite{vgg} with batch normalization layers as our base models on CUB-200 and FoodNet. BCNN is a kind of holistic model for fine-grained classification, which has achieved a good result on CUB-200 dataset. And in the dataset paper of FoodNet \cite{foodnet}, VGG19\_BN has the best performance. Therefore, we choose the above two models as our base models.

\textbf{Implementations.} We implement the base models and inherit most of the hyper parameters from the original papers \cite{bcnn,foodnet}. The SANL blocks are inserted in the BCNN after the last convolution layers of stride 4, 8, and 16. As for VGG19\_BN, the strides are 4, 8, 16 and 32. The spatial attention maps of the SANL blocks are obtained by Grad-CAM, in the same way as our fashion landmark detection model. 

In the training stage, we just resize the image to $448\times448$ for BCNN and $224\times224$ for VGG19\_BN without any data augmentation techniques (crop, flip, etc.). For we like a vanilla version to validate our SANL modules. And that might be the reason that the results of our implementations are slightly different from the results in original papers. 

\textbf{Experimental results.} As shown in Table \ref{tab:fine-grained}, the models with non-local blocks perform even worse than the baseline models on FoodNet and CUB-200, due to a severer over-fitting on small data caused by non-local blocks. While models with the SANL attention mechanism outperform the baseline models by concentrating on the discriminative regions and learning a better representation. 

Our proposed SANL attention is generic and can be plugged into various deep neural network structures. And the experiments on fine-grained classification shows that the SANL attention can improve the performance on the spatial related task other than fashion landmark detection. 

\begin{table}[]
    \centering
    \begin{tabular}{c||c c||c c}
        \multirow{2}{*}{Methods} & \multicolumn{2}{c||}{CUB-200} & \multicolumn{2}{c}{FoodNet} \\
         & top1 & top3 & top1 & top3  \\
        \hline
        Base & 82.15 & 93.12 & 75.14 & 90.13 \\
        NL & 81.57 & 92.60 & 74.63 & 90.31 \\
        SANL & 82.72 & 93.67 & 76.37 & 91.53
    \end{tabular}
    \caption{Experimental results on CUB-200 and FoodNet.}
    \label{tab:fine-grained}
\end{table}

\section{Conclusion}

We have proposed Spatial-Aware Non-Local (SANL) attention mechanism for fashion landmark detection. In the SANL blocks, the spatial attention maps are computed by Grad-CAM on a pretrained network, and fed to the non-local block to help the module concentrate on discriminative regions and learn a spatial related representation. We introduce feature pyramid and coarse-to-fine architecture in our fashion landmark detection framework, and our proposed SANL blocks are plugged into the backbone. We further evaluate our fashion landmark detection approach with the SANL blocks on two large-scale fashion datasets: DeepFashion-C and FLD, and achieve the state-of-the-art performance on both compared with the most recent approaches. The experimental results on fine-grained classification also prove the effectiveness of our proposed SANL attention.


\bibliographystyle{aaai}
\bibliography{aaai2019}
\end{document}